\documentclass[sigconf]{acmart}

\usepackage{graphicx}
\usepackage{caption}
\usepackage{verbatim}
\usepackage{subcaption}
\usepackage{makecell}
\usepackage{multirow} 
\usepackage{colortbl}

\newcommand{\eg}{\emph{e.g.,}}
\newcommand{\etal}{\emph{et al.}}
\usepackage{color}

\AtBeginDocument{%
  \providecommand\BibTeX{{%
    \normalfont B\kern-0.5em{\scshape i\kern-0.25em b}\kern-0.8em\TeX}}}

\begin{document}

\title{Focus on Low-Resolution Information: Multi-Granular Information-Lossless Model for Low-Resolution Human Pose Estimation}


\author{Zejun Gu, Zhong-Qiu Zhao, Hao Shen, Zhao Zhang}


\begin{abstract}
In real-world applications of human pose estimation, low-resolution input images are frequently encountered when the performance of the image acquisition equipment is limited or the shooting distance is too far. However, existing state-of-the-art models for human pose estimation perform poorly on low-resolution images. One key reason is the presence of downsampling layers in these models, \eg~strided convolutions and pooling layers. It further reduces the already insufficient image information.
Another key reason is that the body skeleton and human kinematic information are not fully utilized.
In this work, we propose a Multi-Granular Information-Lossless (MGIL) model to replace the downsampling layers to address the above issues. Specifically, MGIL employs a Fine-grained Lossless Information Extraction (FLIE) module, which can prevent the loss of local information. Furthermore, we design a
Coarse-grained Information Interaction (CII) module  
to adequately leverage human body structural information. 
To efficiently fuse cross-granular information and thoroughly exploit the relationships among keypoints, we further introduce a Multi-Granular Adaptive Fusion (MGAF) mechanism. The mechanism assigns weights to features of different granularities based on the content of the image. The model is effective, flexible, and universal. 
We show its potential in various vision tasks with comprehensive experiments.
It outperforms the SOTA methods by   \textbf{7.7 mAP} on COCO
and performs
well with different input resolutions, different backbones, and different vision tasks.
The code is provided in supplementary material.
\end{abstract}



\maketitle

\section{Introduction}
Human pose estimation aims to localize all human keypoints in a single RGB image. It is the foundation of high-level vision tasks such as action recognition~\cite{vats2022key}, virtual reality~\cite{pham2018human}, and human-computer interaction~\cite{mackenzie2012human}. In recent years, deep learning has greatly improved the performance of human pose estimation.

\begin{figure}\centering
     \flushleft
	\begin{subfigure}{0.15\textwidth}
		\centering
\includegraphics[width=1.0\linewidth,height=1.35\linewidth]{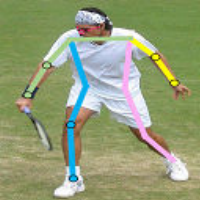}           \caption{128$\times$128}
	\end{subfigure}
	\begin{subfigure}{0.15\textwidth}
		\centering \includegraphics[width=1.0\linewidth,height=1.35\linewidth]{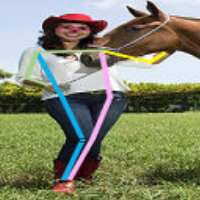}
  \caption{128$\times$128}
	\end{subfigure}
 	\begin{subfigure}{0.15\textwidth}
		\centering \includegraphics[width=1.0\linewidth,height=1.35\linewidth]{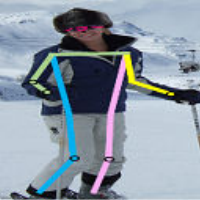}
  \caption{128$\times$128}
	\end{subfigure}
 	\begin{subfigure}{0.15\textwidth}
		\centering \includegraphics[width=1.0\linewidth,height=1.35\linewidth]{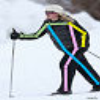}
  \caption{64$\times$64}
	\end{subfigure}
	\begin{subfigure}{0.15\textwidth}
		\centering  
\includegraphics[width=1.0\linewidth,height=1.35\linewidth]{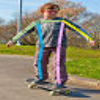}
  \caption{64$\times$64}
	\end{subfigure}
	\begin{subfigure}{0.15\textwidth}
		\centering		\includegraphics[width=1.0\linewidth,height=1.35\linewidth]{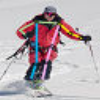}
  \caption{64$\times$64}
	\end{subfigure}

	\caption{Visualization of our proposed MGIL model with low-resolution inputs on the COCO val dataset. }
        
	\label{vis04}
\end{figure}

Recent works~\cite{geng2023human,ye2023distilpose,qu2022heatmap,xu2022vitpose,yang2021transpose,li2021pose,li2021online,kan2023self,yang2023effective,zou2021modulated,zhao2019semantic,zhang2019pose2seg,zeng2020srnet,yang2016end,yang2017learning,xu2021graph,sun2017human,su2019multi,shi2022end,sharma2019monocular} on human pose estimation mainly focus on high-resolution images (greater than or equal to 256$\times$192). However, in real-world application scenarios, it is common to encounter low-resolution input images. For example, due to the large size of high-resolution images, saving or processing them on mobile devices is often challenging. Therefore, low-resolution images have to be used to replace them. Additionally, the limited performance of image acquisition devices or excessive shooting distances may result in low-resolution input images. However, the performance of existing SOTA models for high-resolution human pose estimation significantly degrades when applied to low-resolution images. 
\begin{figure}[htbp]
	\centering	\includegraphics[width=1.00\linewidth,height=0.88\linewidth]{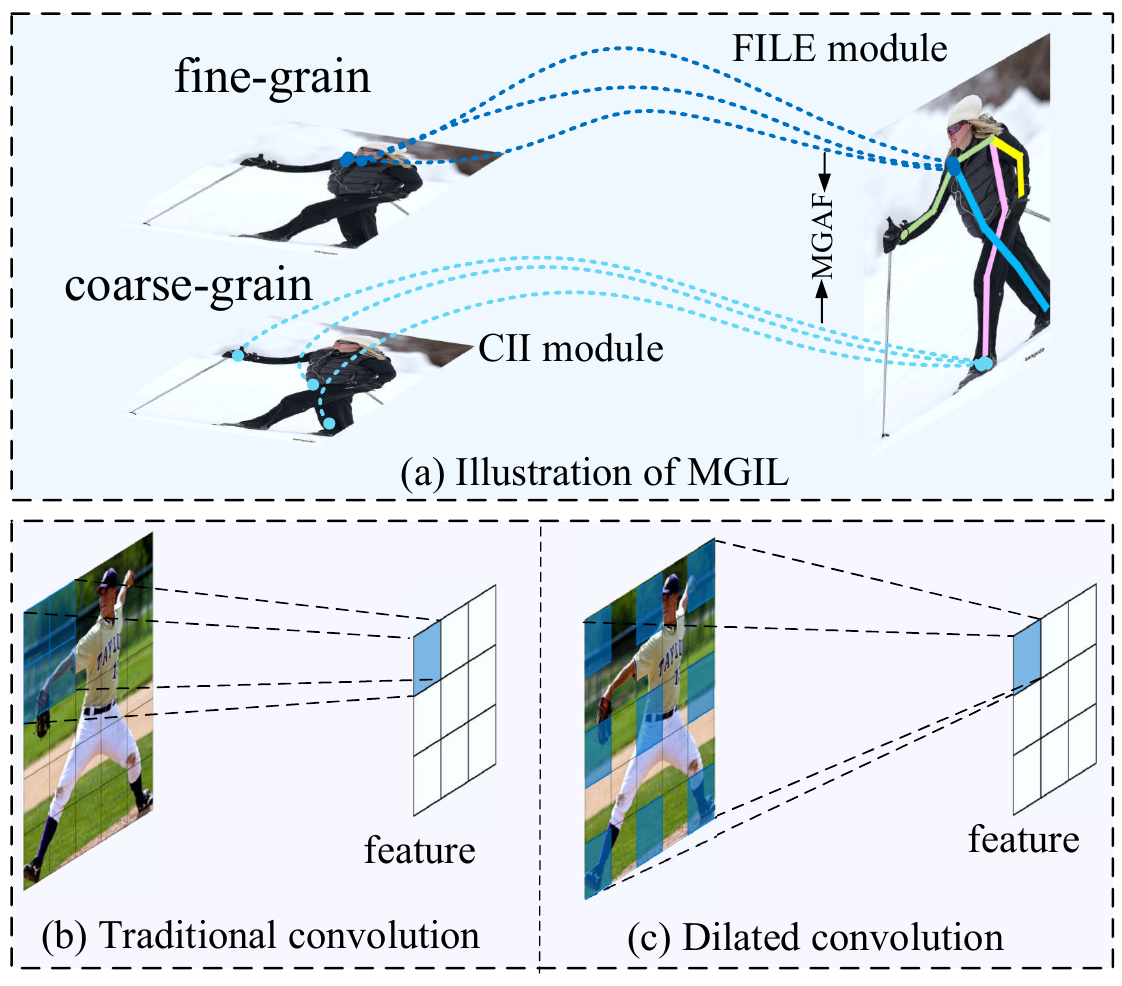}
	\caption{The illustration of  the Multi-Granular
Information-Lossless (MGIL) Model. The comparison of traditional convolution (b) and dilated convolution (c).}
	\label{figillustration}
\end{figure}
Therefore, it is necessary to explore approaches to improve the accuracy of low-resolution human pose estimation.

Current human pose estimation models usually have downsampling layers (strided convolutional layers or pooling layers), such as CPM~\cite{wei2016convolutional}, Hourglass~\cite{newell2016stacked}, and HRNet~\cite{sun2019deep}. These downsampling layers can help models save computational costs and remove redundant information. However, when they are applied to low-resolution images, they will further reduce the already insufficient fine-grained information, which hinders the proper learning of features. In addition, traditional convolutions in their models only focus on local information, neglecting the global information that includes keypoints relationships, as shown in Fig.~\ref{figillustration}~(b). The relational information between keypoints can be leveraged to predict the positions of keypoints based on the human body skeleton structure and human kinematics principles, which is particularly beneficial in conditions of occlusion or blurriness (\eg~low resolution).

To address the above problems, we propose a Multi-Granular Information-Lossless (MGIL) model to replace downsampling layers. 
In the MGIL model, we introduce a Fine-grained Lossless Information Extraction (FLIE) module to tackle the problem of information loss caused by downsampling. 
This module first supplements the information lost due to reduced spatial size onto the channels, achieving a lossless transformation of information, and then ensures the information is fully learned through a Lossless Information Extraction (LIE) block. Furthermore, we introduce a Coarse-grained Information Interaction (CII) module  
to capture information from different receptive fields. The information contains the connectivity and symmetry relationships of the human skeleton which is crucial for keypoint prediction. Finally, we propose a Multi-Granular Adaptive Fusion (MGAF) mechanism to fuse features of different granularities. It can adaptively allocate weights to each feature based on the importance of information at different granularities.

We propose a simple yet sufficiently effective model (MGIL). This model has the following significance: 1) This module ensures that the size of the input and output features remain completely consistent with downsampling layers, thus it can easily replace downsampling layers without any difficulty. 2) This module is versatile, as it can be applied to different convolutional networks, enhancing the performance of various computer vision tasks.


SPD-Conv~\cite{sunkara2022no} is a block that replaces downsampling layers to improve performance on low-resolution visual tasks.  However, it suffers from the following problems: 1) It only utilizes a non-strided convolution, which results in insufficient learning of feature information. 2) It fails to make full use of the spatial correlation information between distant keypoints. Our proposed MGIL model solves the above problems.

Therefore, our contributions can be summarized as:
\begin{itemize}
\item We propose a Multi-Granular Information-Lossless (MGIL) model to replace the downsampling layers to enhance the accuracy of low-resolution human pose estimation. 
\item This model is universal. It can be transferred to various input resolutions, diverse backbones, and
different low-resolution visual tasks (\eg~ classification and object detection).
\item To our best knowledge, we are the first to propose a universal model at the block level to address low-resolution challenges and apply it to human pose estimation.
	\item We show the effectiveness of the MGIL model through extensive experiments. Considerable improvements have been observed in all of these comparisons.
\end{itemize}


%

\section{RELATED WORK}


\subsection{Human Pose Estimation}
In recent years, a substantial amount of work aims to improve the performance of human pose estimation by studying keypoint representation methods, which can be divided into three main directions: heatmap-based methods~\cite{artacho2020unipose,cao2017realtime,fang2017rmpe,li2019rethinking,wang2022lite,xiao2018simple,bulat2016human,tompson2014joint,zhang2020distribution}, regression-based methods~\cite{belagiannis2015robust,carreira2016human,tian2019directpose,toshev2014deeppose,zhou2019objects,varamesh2020mixture}, and other new keypoint representation methods~\cite{li2022simcc,geng2023human}. Heatmap-based methods utilize spatial information around the GT keypoints to create more comprehensive label information. Among the recent heatmap-based studies, some studies~\cite{wei2016convolutional,chen2018cascaded,newell2016stacked,sun2019deep} aim to enhance the model's ability to extract feature information by improving the backbone. Sun \etal~\cite{sun2019deep} adopt a deep high-resolution keypoint representation method, which has become the best-performing backbone in human pose estimation.
Huang \etal~\cite{huang2020devil}
propose to reduce quantization error to improve prediction accuracy. 
Some other works~\cite{li2021tokenpose,xu2022vitpose} utilize transformers to enhance model performance. However, heatmap-based methods have the disadvantage of being inefficient. Regression-based methods are efficient in keypoint prediction by directly regressing the coordinates. Some improved works~\cite{geng2021bottom,wei2020point} propose to focus on local features around the joints. Residual log-likelihood~\cite{li2021human} estimation introduces a new regression paradigm to capture potential output distributions. They all allow regression-based methods to maintain high efficiency. Nevertheless, regression-based methods have the disadvantage of not being accurate enough. The work~\cite{ye2023distilpose} combines the advantages of heatmap-based and regression-based methods by means of knowledge distillation. Recently, several studies focus on new keypoint representations for human pose estimation. 
The works~\cite{li2022simcc,geng2023human} innovatively propose new label representation methods to optimize model learning.

Other studies focus more on practical application issues in human 
pose estimation.  Lee \etal~\cite{lee2023human} introduce ExLPose to address the issue of current models failing to work properly under low-light conditions. 
\begin{figure}[htbp]
	\centering	\includegraphics[width=1.0\linewidth,height=0.72\linewidth]{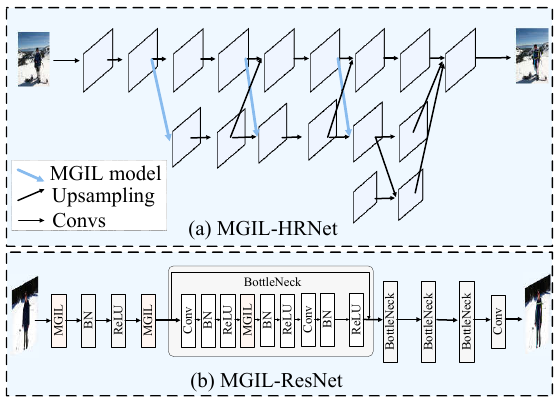}
	\caption{The overall architecture of MGIL-HRNet and  MGIL-ResNet.}
	\label{figbackbone}
 
\end{figure}
Ju \etal~\cite{ju2023human} 
propose a dataset for human pose estimation in artwork to apply the human pose estimation model to virtual scenes. Some works~\cite{yang2023explicit,shi2022end} concentrate on end-to-end multi-person pose estimation, making the model more convenient for practical applications. Li \etal~\cite{yang2023effective} focuses on researching whole-body keypoints to improve the performance of downstream tasks.

\subsection{Low-resolution Vision Tasks}

\begin{figure*}[htbp]
	\centering	\includegraphics[width=0.99\linewidth,height=0.613\linewidth]{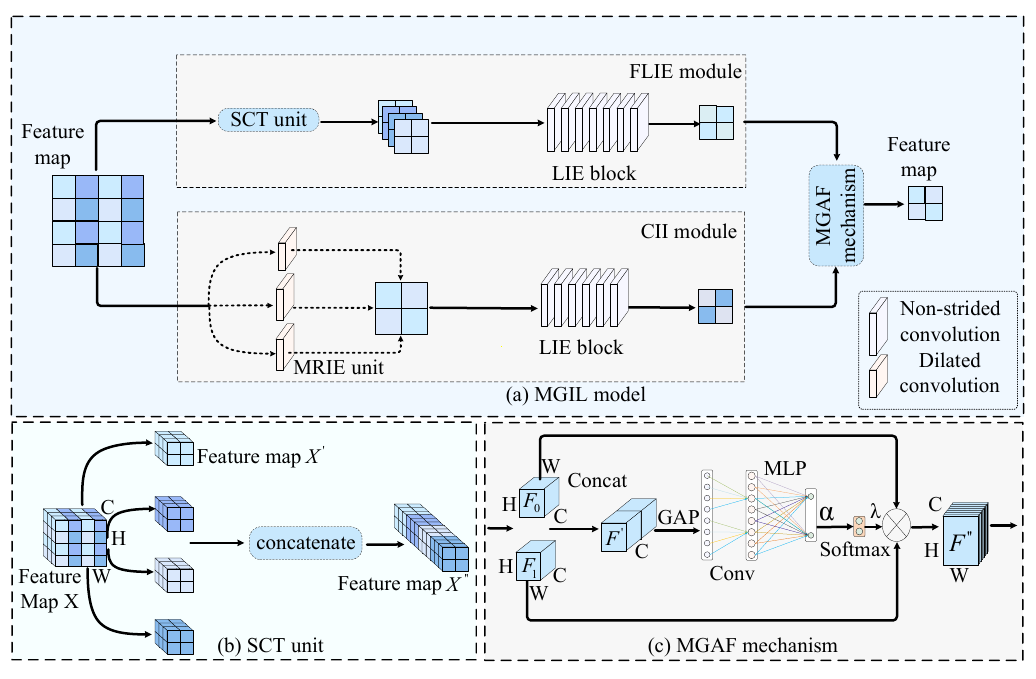}
	\caption{\textbf{(a)} Overview of our proposed  MGIL model. 
   The illustration of SCT unit \textbf{(b)} and MGAF mechanism \textbf{(c)}.
 }
	\label{figframe}
\end{figure*}


Currently, studies on low-resolution images are emerging~\cite{sandler2023parameter,ge2018low,kuen2018stochastic,li2020learning,wang2020resolution,peng2016fine,singh2021enhancing}. Some studies aim at weakening the impact of quantization error in low-resolution instance detection. For example, the work~\cite{wang2022low} minimizes the impact of quantization error on the prediction results by means of confidence-aware learning. The work~\cite{li2022simcc} similarly reduces the quantization error in low-resolution images by transforming the coordinate regression method into a classification task. However, these methods only optimize at the label representation level but do not address the lack of valid information in low-resolution images. Other studies~\cite{shin2022teaching,qi2021multi,huang2022feature} are devoted to further optimizing the low-resolution model by distilling the high-resolution feature information to the low-resolution model. 
Qi \etal~\cite{qi2021multi} propose multi-scale aligned distillation for low-resolution detection. By aligning high-resolution features with low-resolution features, it improves the effect of knowledge distillation. Shin \etal~\cite{shin2022teaching} 
enhance the performance of low-resolution face detection through attention-based knowledge distillation.

There are some studies aiming to improve the backbone for low-resolution detection. The work~\cite{sunkara2022no} proposes SPD-Conv to enhance the performance of low-resolution vision tasks. However, it lacks the utilization of multi-granular information. 
The traditional method~\cite{sun2019deep} of directly increasing image resolution through upsampling interpolation indeed achieves good performance. However, it leads to a significant increase in Params and GFLOPs, making it impractical for performance-limited devices. 
In this work, we propose a Multi-Granular
Information-Lossless (MGIL) model to solve the above problems. Moreover, we are the first to propose a universal model and apply it to low-resolution human pose estimation.

\section{METHODS}
We first introduce the overall framework of our MGIL model and then introduce the designed modules in detail.
\subsection{Overall Framework}\label{sec3.1}
It provides a framework for the proposed model in Fig.~\ref{figframe} (a). The model is composed of a Fine-grained Lossless Information Extraction (FLIE) module, a Coarse-grained Information Interaction (CII) module, and a Multi-Granular 
Adaptive Fusion (MGAF) mechanism. The FLIE module includes a Space-Channel 
Transformation (SCT) unit and a Lossless Information
Extraction (LIE) block. It is introduced in Sec.~\ref{sec3.2}.
 The CII module consists of a Multi-Receptive Information Extract (MRIE) unit and a Lossless Information
Extraction (LIE) block. We describe it in Sec.~\ref{sec3.3}. The MGAF mechanism is described in Sec.~\ref{sec3.4}.

The model we propose can replace the downsampling layer in the current SOTA models, thereby enabling these models to achieve better performance in low-resolution tasks. Fig.~\ref{figbackbone} vividly illustrates the overall architecture of applying the MGIL model to HRNet~\cite{sun2019deep} and ResNet~\cite{he2016deep}.

 \subsection{Fine-grained Lossless Information Extraction (FLIE) Module}\label{sec3.2}

For human pose estimation, cutting-edge CNN models commonly use downsampling layers to reduce redundant information and alleviate computational costs. 
The downsampling layer results in the loss of much fine-grained information in low-resolution images. 
To address this problem, we design a Fine-grained Lossless Information Extraction (FLIE) module.

The operation of traditional downsampling layers can be expressed as:
\begin{equation}
	X^{\prime}=\operatorname{\textit{Down}}(X)
	\label{eq:also-important},
\end{equation}
where $X \in \mathbb{R}^{\mathrm{H} \times \mathrm{W} \times \mathrm{C}}$ denotes the input feature of the downsampling layer, and $\mathit{H}$, $\mathit{W}$, and $\mathit{C}$ are the height, width, and the number of channels of the input feature, respectively; 
$X^{\prime} \in \mathbb{R}^{\mathrm{H}^{\prime} \times \mathrm{W}^{\prime} \times \mathrm{C}^{\prime}}$denotes the output feature of the downsampling layer, and $H^{\prime}$, $W^{\prime}$, $C^{\prime}$  are the height, width, and the number of channels of the output feature maps, respectively; $H^{\prime}$ = $\mathit{H}$/2, $W^{\prime}$ = $\mathit{W}$/2, $C^{\prime}$ = $\mathit{C}$.
The downsampling layer will lead to the loss of feature information. 
Our model needs to preserve all input feature information. 
Meanwhile, to directly replace the downsampling layers and easily apply to various CNN models, MGIL needs to maintain consistency with the input and output feature sizes of the downsampling layer. Thereby, we design a space-channel transformation (SCT) unit.

As depicted in Fig.~\ref{figframe} (b), for any input feature $X \in \mathbb{R}^{\mathrm{H} \times \mathrm{W} \times \mathrm{C}}$ of downsampling layer, we start from different pixel points respectively and sample every two pixels:
\begin{equation}
	X^{\prime}=\operatorname{\textit{Sample}}(X)
	\label{eq:also-important},
\end{equation}
where $\operatorname{\textit{Sample}}(\cdot)$ denotes an interval sampling operation, and $X^{\prime}$ is the output feature of this unit. It can be concretely expressed as:

\begin{equation}
	x_{1}^{\prime} \left.=\left(\begin{array}{ccc}x_{0,0}&x_{0,2}&\cdots\\x_{2,0}&\ddots&\vdots\\\vdots&\cdots&x_{w-2,h-2}\end{array}\right.\right) \\,
\end{equation}
\begin{equation}
	x_{2}^{\prime} \left.=\left(\begin{array}{ccc}x_{1,0}&x_{1,2}&\cdots\\x_{3,0}&\ddots&\vdots\\\vdots&\cdots&x_{w-1,h-2}\end{array}\right.\right) \\,
\end{equation}
\begin{equation}
	x_{3}^{'} \left.=\left(\begin{array}{ccc}x_{0,1}&x_{0,3}&\cdots\\x_{2,1}&\ddots&\vdots\\\vdots&\cdots&x_{w-2,h-1}\end{array}\right.\right) \\,
\end{equation}
\begin{equation}
	x_{4}^{'} \left.=\left(\begin{array}{cccc}x_{1,1}&x_{1,3}&\cdots\\x_{3,1}&\ddots&\vdots\\\vdots&\cdots&x_{w-1,h-1}\end{array}\right.\right),
\end{equation}
where $x_{i}^{\prime} ~$($\mathit{i}$=1,2,3,4) is the sampling result with different coordinate points as starting points, $\mathit{h}$ and $\mathit{w}$ represent the height and width of the feature map, respectively. Then we concatenate $x_{i}^{\prime}$ along the channel dimension after the above process:
\begin{equation}
X^{\prime\prime}=\operatorname{\textit{Concat}}[x_{1,}^{\prime}x_{2,}^{\prime}x_{3,}^{\prime}x_{4}^{\prime}],
\end{equation}
where $\operatorname{\textit{Concat}}[\cdot]$ is the concatenation operator, and $X^{\prime\prime}$ is the result of the concatenation. In this way, while the feature size decreases, the feature information is fully preserved.

After the SCT unit, we further propose a Lossless Information Extraction Module (LIE) block. The LIE block consists of multiple non-strided convolutions. 
On one hand, non-strided convolutions can adequately learn feature information through learnable parameters without any feature information loss. On the other hand, they can also reduce the number of channels to maintain consistency with the modified backbone. 

By combining the SCT unit and the LIE block, fine-grained feature information is fully utilized.  Tab.~\ref{tabeveryunit} provides strong evidence of the performance of the FLIE module.
\subsection{Coarse-grained Information Interaction (CII) Module}\label{sec3.3}
The traditional convolution only focuses on local fine-grained information, as shown in Fig.~\ref{figillustration}~(b), However, 
for human pose estimation, the skeletal structure information, the symmetrical relationship information between joints, and the background information all contribute to inferring joint positions. These all require coarse-grained information from the images, rather than the fine-grained information that traditional convolutions focus on. This work  designs a Coarse-grained Information Interaction
(CII) module to capture coarse-grained information to address the problem.

\begin{table*}[htbp]
  \centering
    \caption{Comparison with SOTA
  methods on the COCO validation dataset.}
  \resizebox{0.99\textwidth}{!}{
  \renewcommand\arraystretch{1.05}
  \setlength{\tabcolsep}{10.0pt}
    \begin{tabular}{cccrrrrrrrr}
    \Xhline{1.3pt}
    Method & Input size & Backbone & AP~$\uparrow$    & AP50~$\uparrow$  & AP75~$\uparrow$  & AP(M)~$\uparrow$ & AP(L)~$\uparrow$ & AR~$\uparrow$    &   \\
    \hline
    HRNet~\cite{sun2019deep} & \multirow{10}[2]{*}{32x32} & HRNet-W32 & 8.2 & 36.9 & 1.3 & 
    9.2 & 6.8 & 15.0   \\
    TokenPose~\cite{li2021tokenpose} &       & HRNet-W32 & 14.0  & 48.2 & 3.4 & 15.2 & 12.5 & 21.7  \\
    CAL~\cite{wang2022low} &       & HRNet-W32 & 26.4 & 61.9 & 18.2 & 27.1 & 25.7 & 33.9  \\
    Rle~\cite{li2021human}&       & HRNet-W32 & 24.4 & -     & -     & -     & -     & -      \\
    Dark~\cite{zhang2020distribution} &       & HRNet-W32 & 12.5 & 45.2 & 2.5 & 13.8 & 11.1 & 20.3  \\
    SimBase~\cite{xiao2018simple} &       & ResNet-152 & 4.4 & 21.1 & 1.0  & 5.3 & 3.2 & 9.0   \\
    PCT~\cite{geng2023human}&       & Swin-Base & 1.3 & 4.6 & 10.0   & 1.0 & 1.2 & 3.1 \\
    Distillpose~\cite{ye2023distilpose} &       & HRNet-W32 & 9.5 & 32.6 & 2.4 & 10.0   & 9.5 & 21.2 \\
  \rowcolor{gray!10}SimCC(\textit{baseline})~\cite{li2022simcc} &       & ResNet-50 & 15.7 & 43.5 & 8.3 & 16.5   & 14.8 & 21.4  \\
    \textbf{MGIL(\textit{ours})} &       & ResNet-50 & \textbf{16.7}\textcolor{blue}{$_{
    +1.0}$}\ & \textbf{45.1} & \textbf{9.4} & \textbf{17.8} & \textbf{15.5} & \textbf{22.6}  \\ 
  \rowcolor{gray!10}SimCC(\textit{baseline})~\cite{li2022simcc} &       & HRNet-W32 & 29.8 & 65.6 & 22.5 & 30.0   & 29.9 & 36.3  \\
    \textbf{MGIL(\textit{ours})} &       & HRNet-W32 & \textbf{37.5}\textcolor{blue}{$_{
    +7.7}$} & \textbf{73.9} & \textbf{34.2} & \textbf{37.5} & \textbf{37.9} & \textbf{43.7}  \\
    \hline
        SimBase~\cite{xiao2018simple} & \multirow{9}[2]{*}{64x64} & ResNet-152 & 30.3 &	67.6&	22.6&	30.6 &	30.5 &	36.2 \\
    Distillpose~\cite{ye2023distilpose} &       & HRNet-W32 & 31.7 & 66.8 & 26.6 & 32.3 & 31.8 & 44.5  \\
    PCT~\cite{geng2023human} &       & Swin-Base  & 11.8 & 37.4 & 4.80 & 12.2 & 12.0 & 17.8  \\
    Rle\cite{li2021human}   &       & HRNet-W32 & 52.5 & -     & -     & -     & -     & -    \\
    HRNet~\cite{sun2019deep} &       & HRNet-W48 & 46.9 & 83.7 & 49.2 & 46.6 & 47.5 & 52.6  \\
    Tokenpose~\cite{li2021tokenpose} &       & HRNet-W48 & 50.3 & 82.7 & 54.4 & 49.9 & 51.4 & 55.6  \\
    CAL~\cite{wang2022low}   &       & HRNet-W48 & 60.6 & 88.1 & 68.4 & 59.5 & 62.3 & 65.5  \\
    Dark~\cite{zhang2020distribution}  &       & HRNet-W48 & 57.2 & 86.8 & 63.5 & 55.9 & 59.2 & 62.2  \\
    \rowcolor{gray!10}Simcc(\textit{baseline})~\cite{li2022simcc} &       & HRNet-W48 & 58.6 & 85.9 & 64.9 & 57.8 & 60.5 & 63.4  \\
    \textbf{MGIL(\textit{ours})} &       & HRNet-W48 & \textbf{63.1}\textcolor{blue}{$_{
    +4.5}$} & \textbf{88.0} & \textbf{70.7} & \textbf{61.6} & \textbf{65.4}  & \textbf{67.5}   \\
    \hline
    \rowcolor{gray!10}SimCC(\textit{baseline})~\cite{li2022simcc} &  \multirow{2}[2]{*}{128x128}      & HRNet-W32 & 72.3 & 91.4 & 80.1 & 70.1   & 76.2 & 75.7  \\
    \textbf{MGIL(\textit{ours})} &       & HRNet-W32 & \textbf{73.0}\textcolor{blue}{$_{
    +0.7}$} & \textbf{91.4} & \textbf{80.2} & \textbf{70.7} & \textbf{76.7} & \textbf{76.1}  \\
    \Xhline{1.3pt}
    \end{tabular}
    }
  \label{tabcoco}%
\end{table*}%

As shown in Fig.~\ref{figillustration}~(c), dilated convolution is the operation of enlarging the convolution kernel by adding gaps between its elements. As the convolutional kernel expands, the corresponding receptive field also increases. Dilated convolution helps to capture contextual information on keypoints.
 We can set different dilation rates to obtain various receptive fields, thus capturing multi-scale information about the human body. This will help the model extract more low-resolution human body information.

In this work, we first adopt dilated convolutions to achieve coarse-grained information extraction and interaction. 
Similar to the FLIE module, to further exploit feature information, we also employ the LIE block within the CII module. 
Table~\ref{tablast} adequately demonstrates the effectiveness of adding the LIE block to the CII module.
The MRIE unit and the CII module complement each other, promoting the utilization of coarse-grained information.

As shown in Fig.~\ref{figillustration}~(a), the FLIE module ensures the lossless transmission of information and fully learns feature information, and the CII module further learns coarse-grained information. 
The combination of the two effectively addresses the issue of severe information deficiency in low-resolution images.

\subsection{Multi-Granular Adaptive Fusion (MGAF) Mechanism}\label{sec3.4}
Usually, the approach to fuse features of different granularities is directly adding them together.
However, for features of different granularities, their importance varies.
Specifically, for occluded keypoints, the local information at the keypoint is not as important as the contextual information. 
Because the local information describes occluding objects rather than keypoints. In contrast, contextual information about keypoints becomes more important.
In this case, coarse-grained features contain more useful information and should be assigned greater weight.

To solve this problem, inspired by the efficient channel attention (ECA) module~\cite{wang2020eca}, we propose a Multi-Granular Adaptive Fusion (MGAF) mechanism.
Fig.~\ref{figframe} (c) illustrates the MGAF mechanism.
 The input for the MGAF mechanism comes from the output feature maps of different granularities.
The MGAF mechanism can assign different weights to features at each granularity based on the corresponding output feature information. 

As shown in Fig.~\ref{figframe} (c), initially, concatenate the two feature maps along the channel dimension and perform global average pooling to obtain 1D contextual features of:

\begin{table*}[htbp]
  \centering
  \caption{Comparison with SOTA
  methods on the MPII val dataset.  }
    \resizebox{0.98
    \textwidth}{!}{
    \renewcommand\arraystretch{1.4}
    \setlength{\tabcolsep}{7.0pt}
    \begin{tabular}{cccrrrrrrrrrr}
    \Xhline{1.3pt}
    Method & Input Size & Backbone  & Head.~$\uparrow$  & Sho.~$\uparrow$  & Elb.~$\uparrow$   & Wri.~$\uparrow$  & Hip.~$\uparrow$   & Knee.~$\uparrow$  & Ank.~$\uparrow$  &  PCKh@0.5~$\uparrow$  \\
    \hline
    SimBase~\cite{xiao2018simple} & \multirow{9}[1]{*}{64x64} & ResNet-152 & 89.905 & 85.258 & 73.854 & 64.058 & 76.216 & 68.628 & 62.659 & 75.353 \\
    OKDHP~\cite{li2021online} &       & 4-Stack HG & 85.573 & 80.757 & 66.099 & 54.328 & 71.767 & 61.032 & 54.393 & 69.037  \\
    HRNet~\cite{sun2019deep} &       & HRNet-W32 & 90.075 & 85.785 & 72.950 & 63.511 & 75.143 & 67.257 & 61.880 & 74.843  \\
    Dark~\cite{zhang2020distribution}  &       & HRNet-W32 & 89.802 & 87.568 & 75.251 & 64.112 & 78.726 & 69.616 & 63.628 & 76.612  \\
    CAL~\cite{wang2022low}   &       & HRNet-W32 & 92.497 & 88.689 & 75.763 & 65.019 & 80.076 & 70.159 & 65.422 & 77.692  \\
    Tokenpose~\cite{li2021tokenpose} &       & HRNet-W32 & 89.018 & 86.770 & 71.570 & 59.963 & 77.237 & 65.947 & 59.139 & 73.997  \\
    PRTR~\cite{li2021pose}  &       & HRNet-W32 & 89.768 & 83.967 & 68.144 & 53.061 & 76.112 & 62.039 & 53.660 & 70.742  \\
    \rowcolor{gray!10}Simcc(\textit{baseline})~\cite{li2022simcc} &       & HRNet-W32 & 92.701 & 88.060 & 76.376 & 66.198 & 77.774 & 68.425 & 64.076 & 77.226 \\
    \textbf{MGIL(\textit{ours})}  &       & HRNet-W32 & \textbf{92.838} & \textbf{89.623} & \textbf{79.461} & \textbf{70.174} & 80.336 & \textbf{72.033} & \textbf{67.524} & \textbf{79.758}\textcolor{blue}{$_{
    +2.532}$} \\
    \hline
    SimBase~\cite{xiao2018simple} & \multirow{8}[2]{*}{32x32} & ResNet-152 & 40.894 & 48.217 & 33.953 & 21.742 & 44.608 & 31.353 & 22.650 & 36.786  \\
    HRNet~\cite{sun2019deep} &       & HRNet-W32 & 46.044  & 52.378  & 40.191  & 28.407  & 48.780  & 37.176  & 27.538  & 42.217   \\
    Dark~\cite{zhang2020distribution}  &       & HRNet-W32 & 38.984 & 61.804 & 46.191 & 31.867 & 61.277 & 45.234 & 28.839 & 48.137  \\
    CAL~\cite{wang2022low}   &       & HRNet-W32 & 77.115 & 68.631 & 48.185 & 33.066 & 63.164 & 46.264 & 40.552 & 55.353  \\
    Tokenpose~\cite{li2021tokenpose} &       & HRNet-W32 & 38.950 & 61.702 & 47.537 & 31.902 & 61.676 & 45.476 & 28.600  & 48.421  \\
    
    PRTR~\cite{li2021pose}  &       & HRNet-W32 & 0.341 & 1.512 & 8.062 & 9.082 & 15.977 & 1.633 & 0.213 & 5.618  \\
    \rowcolor{gray!10}Simcc(\textit{baseline})~\cite{li2022simcc} &       & HRNet-W32 & 81.105  & 72.690  & 54.219  & 37.898  & 63.822  & 51.380  & 46.552  & 59.560  \\
    \textbf{MGIL(\textit{ours})}  &       & HRNet-W32 & \textbf{87.074} & \textbf{79.721} & \textbf{61.633} & \textbf{47.921} & \textbf{69.275} & \textbf{56.297} & \textbf{51.015} & \textbf{65.948\textcolor{blue}{$_{
    +6.388}$}}  \\
    \Xhline{1.3pt}
    \end{tabular}%
    }
    \label{tabmpii}
\end{table*}%

\begin{table}[htbp]
	\centering
	\caption{Performance of our MGIL model in low-resolution object detection tasks on COCO val set.}
    \resizebox{0.48\textwidth}{!}{\renewcommand\arraystretch{1.3}
    \setlength{\tabcolsep}{12.0pt}
	\begin{tabular}{cccc}
		\Xhline{1.1pt}
		Method & Backbone & Image size & AP~$\uparrow$  \\
		\hline
  		\rowcolor{gray!10}YOLOv5n~\cite{jocher2022ultralytics} & YOLOv5 & 640 $\times$ 640 & 28.0 \\
		YOLOv5-SPD-n~\cite{sunkara2022no} & YOLOv5 & 640 $\times$ 640 & 31.0 \\
		YOLOX-Nano~\cite{jocher2022ultralytics} & YOLOv5 & 640 $\times$ 640 & 25.3 \\
  \textbf{MGIL(\textit{ours})}  & YOLOv5 & 640 $\times$ 640 & \textbf{38.7(\textcolor{blue}{$_{
    +10.7}$})}\\
		\Xhline{1.1pt}
	\end{tabular}%
 }
	\label{tabobject}%
\end{table}%

\begin{table}[htbp]
	\centering
	\caption{Performance of our MGIL model in low-resolution image classification tasks.}
 \resizebox{0.49
    \textwidth}{!}{
    \renewcommand\arraystretch{1.3}
	\begin{tabular}{ccc}
		\Xhline{1.3pt}
		Model & Dataset & Top-1 accuracy (\%)~$\uparrow$  \\
		\hline
		\rowcolor{gray!10}ResNet18~\cite{targ2016resnet} & TinyImageNet & 61.68 \\
		Nystromformer~\cite{xiong2021nystromformer} & TinyImageNet & 49.56 \\
		WaveMix-128/7~\cite{jeevan2022wavemix} & TinyImageNet & 52.03 \\
            \textbf{MGIL(\textit{ours})}  & TinyImageNet & \textbf{63.61(\textcolor{blue}{$_{
    +1.93}$})} \\
		\hline
		\rowcolor{gray!10}ResNet50~\cite{koonce2021resnet} & CIFAR-10 & 93.94 \\
		Stochastic Depth~\cite{huang2016deep} & CIFAR-10 & 94.77 \\
		Prodpoly~\cite{chrysos2021deep} & CIFAR-10 & 94.9 \\
            \textbf{MGIL(\textit{ours})}  & CIFAR-10 & \textbf{94.20(\textcolor{blue}{$_{
    +0.26}$})} \\
		\Xhline{1.3pt}
	\end{tabular}%
 }
	\label{tabclass}%
\end{table}%

\begin{table*}[htbp]
	\centering
  	\caption{Ablation studies of each proposed module. Improv. = Improvement. It refers to the improvement of the method relative to the baseline (Simcc~\cite{li2022simcc}). Indivi-improv. = Individual improvement. It refers to the improvement of individual modules compared to the method in the preceding row. The ``Simcc+SPD" method refers to our direct application of the SPD module~\cite{sunkara2022no} to the Simcc~\cite{li2022simcc}. All experiments use HRNet-W32 as the backbone and take images with 32$\times$32 resolution as input.}   
  \resizebox{0.99\textwidth}{!}{
  \renewcommand\arraystretch{1.0}
  \setlength{\tabcolsep}{18.0pt}
	\begin{tabular}{c|ccccccc}
		\Xhline{1.2px}
		Method & FLIE  & CII    & MGAF   & AP~$\uparrow$ &Improv. & Indivi-improv. \\
		\hline
		\rowcolor{gray!10}Simcc~\cite{li2022simcc} & \multicolumn{1}{c}{-} & \multicolumn{1}{c}{-}  & \multicolumn{1}{c}{-} & 29.8 &- & - \\
		Simcc~\cite{li2022simcc}~+~SPD~\cite{sunkara2022no} & \multicolumn{1}{c}{-} & \multicolumn{1}{c}{-}  & \multicolumn{1}{c}{-} & 33.6 & 3.8 & 3.8 \\
		Ours & \checkmark & \multicolumn{1}{c}{-} &  \multicolumn{1}{c}{-} & 36.3 & 6.5 & 6.5 \\
		Ours & \checkmark & \checkmark & \multicolumn{1}{c}{-}  & 36.8 & 7.0 & 0.5 \\
		Ours & \checkmark & \checkmark & \checkmark & 
  37.5 & 7.7 & 0.7 \\
		\Xhline{1.2px}
	\end{tabular}%
 }
	\label{tabeveryunit}%
\end{table*}%

\begin{equation}
F^{^{\prime}}=\frac1{H\times W}\sum_{i=1}^{H}\sum_{j=1}^{W}[F_0,F_1](i,j), 
\end{equation}
where $F_0, F_1$ are the output feature maps of each granularity respectively, $\begin{bmatrix}\cdot\end{bmatrix}$ indicates the concatenate operation of feature maps along the channel dimension, and    $\mathit{H}$, $\mathit{W}$  are the height and width of the feature map respectively. The global pooling operations 
are advantageous for capturing spatial global information of features, which thus aids in the fusion of different features. Then  $F^{\prime}$ is fed into a one-dimensional convolution and then into a fully connected layer to generate the weight matrix:

\begin{equation}
\mathbf{\alpha}=MLP(Conv(F^{\prime})),
\end{equation}
where $\operatorname{\textit{Conv}}$ is an 1x1 convolution with an adaptive number of output channels. $\operatorname{\textit{MLP}(\cdot)}$ consists of a ReLU function and an FC layer with 2 output channels. The convolution operation can fully learn the correlation between the feature channels. The $\operatorname{\textit{MLP}}$ operator serves two purposes: firstly, they reduce the number of feature channels to match the number of granularities. Secondly, they complement convolution operations, further enhancing the learning of channel correlations, and thereby optimizing weight allocation. Finally, a Softmax operation is applied to the weight matrix to obtain the weight coefficients for different granularities:
\begin{equation}   \lambda=\operatorname{\textit{Softmax}}(\alpha),
\end{equation}
 where $\lambda$ is a sequence of 2-dimensional weights. The features of different granularities are combined with their corresponding weight to obtain the final fusion result:
\begin{equation}
F^{\prime\prime}={\lambda}_{0}\cdot F_{0}+{\lambda}_{1}\cdot F_{1},
\end{equation}
where ${\lambda}_{0}$ and ${\lambda}_{1}$ are the coefficients corresponding to each granularity, respectively, and  $F^{\prime\prime}$ are the fused features. $F^{\prime\prime}$ are then fed into the subsequent network structure. 
The MGAF mechanism can learn spatial global information and channel correlations from each granularity, adaptively assigning weights to them, and ultimately producing a more effective output.

\section{EXPERIMENTS}
We first evaluate the proposed MGIL model on two benchmark datasets for human pose estimation: COCO~\cite{lin2014microsoft} and MPII~\cite{andriluka20142d}.
Then we present the results of object detection and classification.
Ablation studies about the main components of MGIL are also provided to help understand the approach.

\begin{table}[htbp]
	\centering
	\caption{Ablation studies of the number of non-strided convolution in the LIE block. ``Conv number" refers to the total count of non-strided convolutions within the LIE block of the FILE module.
	}
  \resizebox{0.47
    \textwidth}{!}{
    \renewcommand\arraystretch{1.3}
	\setlength{\tabcolsep}{6.5pt}
	\begin{tabular}{ccccc}
		\Xhline{1.3pt}
		Method & Conv number& Backbone & Input size & AP~$\uparrow$ \\
		\hline
		\rowcolor{gray!10}Simcc~\cite{li2022simcc} & - & HRNet-W32 & 32x32 & 29.8  \\
		Ours  & 1     & HRNet-W32 & 32$\times$32 & 33.6  \\
		Ours  & 2     & HRNet-W32 & 32$\times$32 & 36.3  \\
		Ours  & 3     & HRNet-W32 & 32$\times$32 & \textbf{~36.6 } \\
		\Xhline{1.3pt}
	\end{tabular}%
 }
	\label{tabconvnumber}%
\end{table}%

\subsection{Datasets and Metrics}
\textbf{Human pose estimation datasets.} To validate the effectiveness of our proposed MGIL model, we conduct experiments on two popular human pose datasets: COCO~\cite{lin2014microsoft} and MPII~\cite{andriluka20142d}. 
As one of the largest and most challenging datasets for human pose estimation, the COCO dataset contains more than 200K images and 250K human instances labeled with 17 keypoints. The COCO dataset is divided into three parts:57k images for the training set, 5k images for val set, and 20k images for the test-dev set. In this paper, we follow the data expansion in~\cite{sun2019deep}.
The MPII dataset contains approximately 25K
images including over 40K subjects with annotated body joints, where 29K subjects are used for training and 11K subjects are used for testing. The images are collected using an established taxonomy of everyday human activities from YouTube videos. We adopt the same train/valid/test split as in~\cite{zhang2019fast}. In the MPII dataset, each person instance has 16 labeled joints.

\textbf{Object detection datasets.} To evaluate the performance of our model on the object detection task, we adopt the COCO~\cite{lin2014microsoft} dataset. 

\textbf{Classification datasets.} The CIFAR-100~\cite{krizhevsky2010cifar} dataset consists of colored natural images with 32 $\times$ 32 pixels. The training and testing
sets contain 50K and 10K images, respectively. 
\begin{figure*}
	\centering
	\begin{subfigure}{0.16\textwidth}
		\centering		\includegraphics[width=1.0\linewidth,height=1.5in]{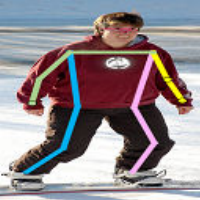}
	\end{subfigure}
	\begin{subfigure}{0.16\textwidth}
		\centering
		\includegraphics[width=1.0\linewidth,height=1.5in]{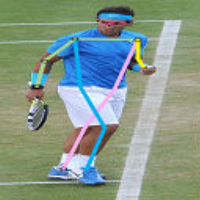}
	\end{subfigure}
	\begin{subfigure}{0.16\textwidth}
		\centering
		\includegraphics[width=1.0\linewidth,height=1.5in]{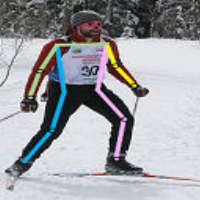}
	\end{subfigure}
	\begin{subfigure}{0.16\textwidth}
		\centering
		\includegraphics[width=1.0\linewidth,height=1.5in]{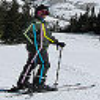}
	\end{subfigure}
	\begin{subfigure}{0.16\textwidth}
		\centering
		\includegraphics[width=1.0\linewidth,height=1.5in]{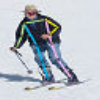}
	\end{subfigure}
    \begin{subfigure}{0.14\textwidth}
        \centering
        \includegraphics[width=1.0\linewidth,height=1.5in]{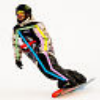}
    \end{subfigure}
	\caption{Visual results of our model under low-resolution conditions based on the COCO val dataset. From left to right, the resolutions are 128$\times$128, 128$\times$128, 128$\times$128, 64$\times$64, 64$\times$64, and 64$\times$64, respectively.}
	\label{figvis_2}
\end{figure*}
TinyImageNet~\cite{le2015tiny} is a subset of the
ILSVRC-2012 classification dataset and contains 200 classes. Each class has
500 training images, 50 validation images, and 50 test images. Each image is of resolution 64$\times$64$\times$3 pixels.

\textbf{Evaluation metrics.}
For human pose estimation, we follow the standard evaluation metrics for the COCO~\cite{lin2014microsoft} and MPII~\cite{andriluka20142d}. In particular, 
OKS defines the similarity between
different human poses. 
The OKS-based AP (average precision) is reported for the COCO dataset. 
The PCK metric refers to the percentage of correct keypoints that lie within a normalized distance of ground truth. 
We adopt the PCKh@0.5  as
our evaluation metric  for  the MPII dataset,
which entails a threshold of 50\% of the head diameter.

For object detection, we 
report the standard metric of average precision (AP) on val2017 under different
IoU thresholds [0.5:0.95] and object sizes (small, medium, large). 
For classification, we use top-1 accuracy as the metric to evaluate classification performance.

\subsection{Implementation Details}

For all methods used to compare, we replicate them following their original settings in corresponding papers and code. We only modify the image input resolution, while keeping all other settings unchanged. \textbf{\textit{For all methods, we train and test under the same low resolution.}} We adopt the top-down estimation pipeline. 
We use a commonly used person detector provided by SimpleBaselines~\cite{xiao2018simple} with 56.4 AP for the COCO val dataset.

We first train the SOTA models at low resolution and then test them at the same resolution.
 Experimental results indicate that the Simcc~\cite{li2022simcc} method achieves the best performance. We adopt it as our base model.
The training settings for our method, such as the optimizer,
learning rate, and data augmentation, are the same as Simcc~\cite{li2022simcc}. All experiments are implemented using the PyTorch library on two NVIDIA GeForce 3080-Ti GPUs.
\subsection{Results in Human Pose Estimation}
\textbf{COCO.} We replicate the existing SOTA models at different low resolutions. 
We evaluate our methods on the COCO validation dataset. The experimental results are shown in Tab.~\ref{tabcoco}. We can
see that MGIL achieves the best performance compared with other SOTA methods. In particular, we surpass the baseline method (Simcc~\cite{li2022simcc}) by 7.7 mAP. Our method achieves a consistent performance improvement in different resolutions and backbones. 
It fully demonstrates that our proposed MGIL is an effective and universal model for low-resolution human pose estimation.

\textbf{MPII.}  The results on the MPII validation set are shown
in Tab.~\ref{tabmpii}. We also conduct experiments at different resolutions.
Our approach significantly surpasses the other
methods. Compared to the baseline
method SimCC~\cite{li2022simcc}, our method achieves an improvement
of 6.388 under the metric of PCKh@0.5 when the input resolution is 32$\times$ 32. It further validates the effectiveness of our approach.

\subsection{Results on Other Computer Vision Tasks}

\textbf{Object detection.} To demonstrate the generality of our approach, we apply the model to low-resolution object detection tasks and conduct experiments on the COCO dataset, as shown in Tab.~\ref{tabobject}. We replace the downsampling layer in YOLOv5n with MGIL. Our method outperforms the baseline method by 10.7 mAP. 
The effectiveness and generality of our MGIL are thoroughly validated.

\begin{table}[htbp]
	\centering
	\caption{Ablation studies of the LIE block in different modules. ``Conv number" refers to the sum of non-strided convolutions.}
   \resizebox{0.47
    \textwidth}{!}{
    \renewcommand\arraystretch{1.1}
	\setlength{\tabcolsep}{6.5pt}
	\begin{tabular}{ccccc}
		\Xhline{1.3pt}
		Method & Conv number in FILE & Conv number in CII  & AP  \\
		\hline
		Simcc~\cite{li2022simcc} & - & -  & 29.8  \\
		\hline
		\rowcolor{gray!10}Ours  & 1     & -      & 33.6  \\
		Ours  & 1     & 1          & 34.6  \\
		\hline
		\rowcolor{gray!10}Ours  & 2     & -  & 36.3  \\
		Ours  & 2     & 1  & 36.8  \\
		Ours  & 2     & 2  & 37.1  \\
		\Xhline{1.3pt}
	\end{tabular}%
	\label{tablast}%
 }
\end{table}%

\textbf{Classification.} In this section, we verify the performance of our MGIL model in low-resolution classification. The experiments are conducted on two datasets: CIFAR-10 and TinyImageNet. Similarly, we replace the downsampling layer with MGIL. As shown in Tab.~\ref{tabclass}, our method improves the performance of both datasets showing that our method is versatile for low-resolution classification tasks.

\subsection{Ablation Studies}
\textbf{Each component.} In this subsection, we conduct ablation experiments to demonstrate the contribution of each proposed component. As shown in Tab.~\ref{tabeveryunit}, all the proposed components contribute to the improvement of the model's performance. Compared to traditional downsampling layers, the ``Simcc+SPD" method improves performance by 3.8 mAP. Our proposed Fine-grained Lossless Information Extraction (FLIE) module, Coarse-grained Information Interaction (CII)
module  and Multi-Granular Adaptive Fusion (MGAF)
mechanism improves performance by 6.5, 0.5, and 0.7 mAP, respectively.
The ablation experiments demonstrate the complementary roles of each of our proposed modules, and the combination of all proposed methods results in the best performance, with a significant 7.7 mAP improvement over the baseline.

\textbf{Conv number in the LIE block.}
To verify the impact of the number of non-stride convolutions within the LIE block on the module's performance,  we conduct experiments with various numbers of non-stride convolutions. As shown in Tab.~\ref{tabconvnumber}, as the number of non-strided convolutions increases, the model's accuracy gradually improves. 
\begin{table}[htbp]
	\centering
	\caption{Ablation studies for the Multi-Granular Adaptive Fusion (MGAF)
mechanism.  ``Conv number" refers to the sum of non-strided convolutions.
	}
   \resizebox{0.47
    \textwidth}{!}{
    \renewcommand\arraystretch{1.5}
	\setlength{\tabcolsep}{6.5pt}
	\begin{tabular}{cccccc}
		\Xhline{1.3pt}
		Method & Conv number in FILE & Conv number in CII & AF   & AP~$\uparrow$ \\
		\hline
		\rowcolor{gray!10}Ours  & 1     & 1     & ×     & 34.6  \\
		Ours  & 1     & 1     & \checkmark     & 35.7 \\
		\hline
		\rowcolor{gray!10}Ours  & 2     & 2     & ×     & 37.1  \\
		Ours  & 2     & 2     &\checkmark     & 37.5  \\
		\Xhline{1.3pt}
	\end{tabular}%
 }
	\label{tabMGAF}%
\end{table}%
It indicates that increasing the number of non-strided convolutions allows for further learning of low-resolution information. 
However, excessive non-strided convolutions also increase the model's computational cost. Therefore, we choose 3 as the number of convolutions in the LIE block.

\textbf{ LIE block in different modules.}
 The LIE block is composed of several non-strided convolutions.
Tab.~\ref{tablast} shows the number of non-strided convolutions in different modules.
The results indicate that the LIE block brings substantial improvement in both the FILE and CII modules.

\textbf{MGAF mechanism.}
We perform ablation experiments of the  Multi-Granular Adaptive Fusion
(MGAF) mechanism in different model structures. We realize different model structures by setting the various number of non-stride convolutions in each module. As shown in Tab.~\ref{tabMGAF}, relative to the direct additive fusion method, the MGAF mechanism improves 1.1 mAP and 0.4 mAP, respectively. It proves that the MGAF mechanism can further extract effective feature information. It has excellent performance and robustness.

\subsection{Visualized Results}
Fig.~\ref{figvis_2} provides more visualized human pose estimation results. As is shown in Fig.~\ref{figvis_2}, the MGIL model achieves accurate human pose estimation when the input is low-resolution.

\section{CONCLUSION}
In this work, we propose a novel Multi-Granular Information-Lossless (MGIL) model, which can significantly improve the performance of low-resolution vision tasks. Our proposed model adaptively extracts multi-granular feature information, greatly supplementing the effective feature information that originally was not sufficient under low-resolution conditions. Our work truly discovers and addresses the pain points of low-resolution visual tasks.
Numerous experiments show its versatility and effectiveness. 
It can easily replace the downsampling layer in various backbones, various input resolutions, and various low-resolution vision tasks. 
Next, we plan to extend MGIL to large models to achieve greater improvements in performance on low-resolution vision tasks.

\newpage

\bibliographystyle{ACM-Reference-Format}
\bibliography{sample-base}


\end{document}